# Real-Time Surface-to-Air Missile Engagement Zone Prediction Using Simulation and Machine Learning

**Joao P. A. Dantas, Diego Geraldo, Felipe L. L. Medeiros**

**Institute for Advanced Studies**

**Sao Jose dos Campos, SP, Brazil**

dantasjpad@fab.mil.br
diegodg@fab.mil.br
felipefllm@fab.mil.br

**Marcos R. O. A. Maximo, Takashi Yoneyama**

**Aeronautics Institute of Technology**

**Sao Jose dos Campos, SP, Brazil**

mmaximo@ita.br
takashi@ita.br

## ABSTRACT

Surface-to-Air Missiles (SAMs) are crucial in modern air defense systems. A critical aspect of their effectiveness is the Engagement Zone (EZ), the spatial region within which a SAM can effectively engage and neutralize a target. Notably, the EZ is intrinsically related to the missile's maximum range; it defines the furthest distance at which a missile can intercept a target. The accurate computation of this EZ is essential but challenging due to the dynamic and complex factors involved, which often lead to high computational costs and extended processing times when using conventional simulation methods. In light of these challenges, our study investigates the potential of machine learning techniques, proposing an approach that integrates machine learning with a custom-designed simulation tool to train supervised algorithms. We leverage a comprehensive dataset of pre-computed SAM EZ simulations, enabling our model to accurately predict the SAM EZ for new input parameters. It accelerates SAM EZ simulations, enhances air defense strategic planning, and provides real-time insights, improving SAM system performance. The study also includes a comparative analysis of machine learning algorithms, illuminating their capabilities and performance metrics and suggesting areas for future research, highlighting the transformative potential of machine learning in SAM EZ simulations.

## ABOUT THE AUTHORS

**João P. A. Dantas** received his B.Sc. degree in Mechanical-Aeronautical Engineering from the Aeronautics Institute of Technology (ITA) in Brazil in 2015. During this time, he participated in a year-long exchange program at Stony Brook University (SBU) in the USA. He also obtained his M.Sc. degree from ITA's Graduate Program in Electronic and Computer Engineering in 2019. In 2022, he worked as a visiting researcher in the AirLab at the Robotics Institute of Carnegie Mellon University. He is pursuing his Ph.D. in the same program at ITA and working as a researcher for the Brazilian Air Force at the Institute for Advanced Studies (IEAv). His research interests include Artificial Intelligence, Machine Learning, Robotics, and Simulation.

**Diego Geraldo** was commissioned in 2005 as a Brazilian Air Force Academy graduate. In 2006, he completed his operational specialization as a fighter pilot. He received his M.Sc. degree from the Graduate Program in Aeronautical and Mechanical Engineering at the Aeronautics Institute of Technology (ITA) in 2012. Since then, he has been a researcher for the Brazilian Air Force (FAB) at the Institute for Advanced Studies (IEAv), focusing on Modeling and Simulation, Artificial Intelligence, and Data Science while acting as a developer of Decision Support Systems for FAB.

**Felipe L. L. Medeiros** received a bachelor's degree in Computer Science (1999) from the Federal University of Ouro Preto, Brazil, and the M.Sc (2002) and Ph.D. (2012) degrees in Applied Computing from the National Institute for Space Research, Brazil. He has been working in artificial intelligence, emphasizing metaheuristics, machine learning, autonomous agents, and simulation.

**Marcos R. O. A. Maximo** received the B.Sc. degree (Hons.) (Summa cum laude) in computer engineering and the M.Sc. and Ph.D. degrees in electronic and computer engineering from the Aeronautics Institute of Technology (ITA),

Brazil, in 2012, 2015, and 2017, respectively. He is currently a Professor at ITA, an Autonomous Computational Systems Laboratory (LAB-SCA) member, and leads the Robotics Competition Team: ITAndroids. He is especially interested in humanoid robotics. Moreover, his research interests include mobile robotics, dynamical systems control, and artificial intelligence.

**Takashi Yoneyama** received a bachelor's degree in electronic engineering from the Aeronautics Institute of Technology (ITA), Sao Jose dos Campos, Brazil, in 1975, an MD degree in medicine from the Taubate University, Taubate, Brazil, in 1993, and the Ph.D. degree in electrical engineering from the Imperial College London, London, U.K., in 1983. He is a Professor of Control Theory with the Department of Electronics, ITA. He has more than 300 published papers, has written four books, and supervised more than 70 theses. His research is concerned mainly with stochastic optimal control theory. Prof. Yoneyama was the President of the Brazilian Automatics Society from 2004 to 2006.

# Real-Time Surface-to-Air Missile Engagement Zone Prediction Using Simulation and Machine Learning

**Joao P. A. Dantas, Diego Geraldo, Felipe L. L. Medeiros**

**Institute for Advanced Studies**

**Sao Jose dos Campos, SP, Brazil**

dantasjpad@fab.mil.br
diegodg@fab.mil.br
felipefllm@fab.mil.br

**Marcos R. O. A. Maximo, Takashi Yoneyama**

**Aeronautics Institute of Technology**

**Sao Jose dos Campos, SP, Brazil**

mmaximo@ita.br
takashi@ita.br

## INTRODUCTION

Surface-to-Air Missiles (SAMs) are indispensable in modern air defense systems, forming a robust line of defense against airborne threats (Open'ko et al., 2017). The operational effectiveness of these missile systems depends heavily on accurately determining their Engagement Zone (EZ) — the spatial region in which a SAM can successfully engage and potentially eliminate a target (Leonard, 2011). However, this parameter's complexity is influenced by many factors, including the missile's propulsion and guidance systems, the target's unique characteristics, such as speed, altitude, off-boresight angle, and the evasive maneuver pattern (Alberts et al., 2000). The dynamic nature of these factors necessitates robust, efficient simulation tools capable of accurately predicting the EZ, a critical factor in formulating military strategies (Dantas et al., 2021a).

Nevertheless, conventional simulation methods for estimating missile EZs have been associated with high computational costs and lengthy processing times. Researchers have consistently pointed out that these limitations can significantly interfere with the speed and effectiveness of defense strategies (Birkmire, 2011). Given these challenges, this study explores the potential of machine learning techniques as an alternative approach to address these problems.

By leveraging machine learning algorithms, known for their effectiveness in optimizing complex computational processes (Sun et al., 2019), this study aims to aid the computational limitation traditionally associated with SAM EZ simulations. Using a comprehensive dataset of pre-computed SAM EZ simulations, our proposed method integrates machine learning with a custom-designed simulation tool to train supervised algorithms – renowned for effectively addressing such issues (Dantas et al., 2022a). This approach, inspired by Dantas et al. (2021a), uses the extensive training phase to enable our model to accurately predict the EZ of a SAM for a new set of input parameters.

Despite the promising potential of this approach in addressing defense-related issues, it also poses distinct challenges. Specifically, when dealing with missile launch data, one has to navigate through obstacles such as the need for an extensive amount of training data, the risk of model overfitting, and the perpetual necessity for meticulous evaluation and refinement to assure model precision and applicability (Dantas, 2018; Dantas et al., 2021b; Dantas et al., 2022b). Nonetheless, these challenges also provide a route for further research and improvement, underscoring the continual need for innovation in this area.

The main contribution of this work is to highlight the transformative potential of machine learning techniques to speed up the response to SAM EZ simulations. It can contribute significantly to air defense strategy planning and potentially enhance SAM system performance by offering real-time insights. In doing so, we aim to combine the strengths of simulation and machine learning to bring a new level of dynamism and responsiveness to military strategy, enabling more informed and timely decision-making. Furthermore, our work includes a comparative analysis of widely recognized machine learning algorithms, emphasizing their performance metrics, training, and inference durations. This comparison offers a better understanding of these algorithms' capabilities and areas necessitating enhancement.

The structure of this paper extends as follows: The subsequent section delves into a comprehensive review of the related work, setting the context for the study. This is followed by a detailed description of our research's simulation and analysis tools. Subsequently, we elucidate the methodology for developing machine learning algorithms to calculate the SAM EZ. After that, we present the results of our investigation and engage in a thorough discussion of their implications. Finally, we draw our conclusions and propose directions for future research.

**RELATED WORK**

The simulation of missile-target engagements, essential for evaluating and developing defense systems, has been a research focus for several decades. Traditional simulation methods have been noted for their accuracy but are often constrained by high computational costs and lengthy processing times. This has spurred the investigation of alternative approaches, including machine learning techniques, to optimize efficiency.

Mathematical models for missile-target engagements are among this domain's foundational literature. For instance, Philips (1991) presented a seminal work in which the Engagement Envelope Generator (EEG) was developed. The EEG, designed modularly, could be integrated with various mid-course missile trajectory simulations. The Rapid Estimation of Terminal Performance (RETP) algorithm, which utilizes linear adjoint techniques, was developed to rapidly estimate the miss distance and probability of hit. Philips' multilevel approach suggests the use of post-processing Monte-Carlo simulations for improved accuracy in analyzing intercepts along the EEG-generated engagement envelopes' boundaries.

Building on the earlier mathematical models, Farlik et al. (2017) investigated military modeling and simulation, emphasizing troops' training and military scenario verification. They highlighted the partial development of simulations for ground-based air defense systems and discussed an approach for modeling missile systems' firing capabilities. This involved simplifying missile systems' effective coverage zones for seamless integration into military simulators.

Furthermore, Li et al. (2020) developed a refined model for air-to-air missile attack zones by incorporating off-axis angles, allowing missiles to hit targets from different angles. They used the dichotomy search and fourth-order Runge-Kutta methods to calculate missile trajectories and determine hit conditions. Their results showed variations in the attack zone boundaries based on the target entry and missile off-axis angles, providing valuable insights for pilots in deciding whether to launch missiles.

Recent studies have made headway in employing machine learning techniques, particularly neural networks, to predict the Weapon Engagement Zone (WEZ). Yoon et al. (2010) used a Wavelet Neural Network to improve the accuracy of the weapon launch acceptability region, with promising results in terms of both accuracy and memory requirement reductions.

Similarly, Birkmire (2011) investigated a Multilayer Perceptron (MLP) with Bayesian Regularization to approximate the WEZ's maximum launch range. The network was trained on simulated data and showed improved approximation accuracy, demonstrating the feasibility of integrating ANNs into practical models.

Additionally, Dantas et al. (2021a) demonstrated using a Deep Neural Network (DNN) to estimate the WEZ's maximum launch range. By training the DNN on simulated launches, the model could efficiently predict the WEZ under various firing conditions. This study illustrated the synergy of simulation and machine learning in improving computational efficiency in missile-target engagement predictions.

In summary, the existing literature demonstrates a progression from mathematical models to the application of machine learning techniques in missile-target engagement simulations. This evolution highlights the trade-off between the accuracy of models and computational efficiency. Our research takes inspiration from the work of Dantas et al. (2021a) to integrate simulation with machine learning, focusing on the real-time prediction of SAM EZ under various firing conditions. In contrast to the aforementioned study, which focused on air-to-air missiles and analyzed missile models for a broad interval between 0 and 180 degrees, our work is tailored to SAM systems and conducts a more granular analysis by dividing models into sectors based on aspect angles. Additionally, we systematically compare three supervised machine learning algorithms in terms of performance and time response, as opposed to the sole use of a DNN in the prior study. This multifaceted approach enables us to address the unique challenges associated with SAM systems.

**SIMULATION AND ANALYSIS TOOLS**

This study employs a simulation tool adapted from the model proposed by Dantas et al. (2021a), modified to model Surface-to-Air Missile (SAM) dynamics rather than air-air missiles. Constructed in the R programming language, the tool effectively leverages R's robust statistical capabilities and advanced graphical functions, making it suitable for simulating SAM dynamics (Ihaka & Gentleman, 1996). Our tool features an intricately designed 5-degree-of-freedom (5DOF) model to accurately simulate a Fox 3 missile – an active radar-guided missile with an autonomous target-tracking seeker post-activation (US Department of Defense, 1995).

Our tool meticulously emulates this guidance functionality, facilitating perfect proportional navigation alongside an aggressive post-launch climb—commonly known as a loft maneuver (US Department of Defense, 1995). Its adaptive capacity allows for the simulation of missile trajectories considering both stationary and maneuvering targets, thereby comprehensively exploring diverse engagement scenarios. These characteristics are vividly illustrated in Figure 1.

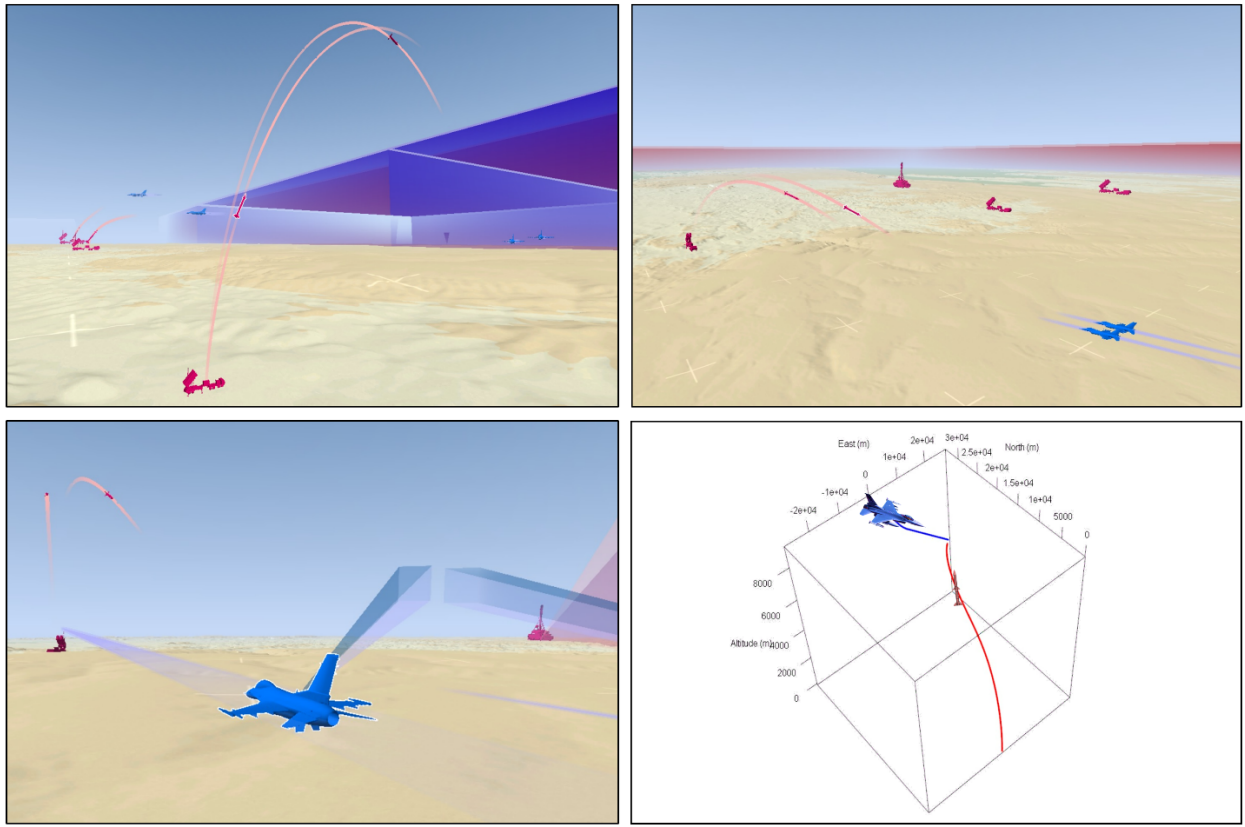

**Figure 1. Illustration of a red simulated missile engagement against a blue target.**

Representing SAM dynamics with high fidelity, this tool extends beyond numerical experiments to offer a scalable platform for creating, testing, and refining SAM models. Its potential for further expansion includes incorporating complex scenarios, diverse missile types, and environmental factors, making it an invaluable asset in EZ prediction and broader military strategy development.

We use the Aerospace Simulation Environment (*Ambiente de Simulação Aeroespacial – ASA* in Portuguese) to enhance launch visualization and debugging processes. This high-fidelity C++ simulation framework, developed specifically for the Brazilian Air Force (Dantas et al., 2022c), facilitates the modeling and simulation of diverse military operational scenarios. For the analysis of simulation data, we employ AsaPy (Dantas et al., 2023), a custom-made Python library associated with ASA, specifically designed to optimize the analysis of simulation data and is an integral component of the ASA technology. This library augments our data preprocessing, feature extraction, and machine learning algorithm training and evaluation stages, improving the overall efficiency of our study.

**METHODOLOGY**

In this study, we utilized a specialized simulation tool, as previously discussed, to estimate the maximum range of two distinct types of SAMs, which we have labeled SAM$_1$ and SAM$_2$. It is important to note that the actual models of these SAMs are not disclosed due to the sensitive nature of defense information. SAM$_1$ and SAM$_2$ are engineered for distinct operational requirements. Specifically, SAM$_2$ is optimally designed for long-range engagements and is especially effective against ballistic missiles, owing to its extended range capabilities. In contrast, SAM$_1$ excels in medium-range engagements, with high maneuverability and precision targeting offering distinct advantages. This differentiation uniquely suits each missile system to address varied threat profiles and operational contexts.

It is crucial to note that the maximum engagement range of a SAM, which is contingent on various situations between the missile launcher and the target, characterizes its EZ. For the models proposed in this study, we consider three main parameters as input:

- The elevation of the target relative to the launcher ($x_1$), spanning a range from -5,000 ft to 45,000 ft;
- The speed of the target ($x_2$), measured in knots, encompassing values from 200 kt to 850 kt; and
- The absolute value of the aspect angle of the target in relation to the launcher ($x_3$), varying within a spectrum from 0 degrees to 180 degrees.

Our tool leverages these parameters to calculate a missile's maximum range (y) when launched toward a target. In this particular scenario, in each simulation run, the target is characterized as a passive aircraft, which maintains a state of non-reactivity during its flight - sustaining a constant speed and unwavering altitude throughout.

Initially, we generated 3,000 samples of the maximum range for each SAM type. Following their intervals, this was accomplished by applying Latin Hypercube Sampling (LHS) to the independent variables $x_1$, $x_2$, and $x_3$. Each resultant sample, referred to as an observed sample, comprises the values of the independent variables $x_1$, $x_2$, and $x_3$, along with a corresponding value of the dependent variable y.

Subsequently, we employed a simple polynomial regression to construct EZ models for SAM$_1$ and SAM$_2$ based on the generated sample sets. These models enable predictions of each SAM type's performance under varying conditions, as dictated by the input parameters.

We accomplished a set of tests with the EZ models created and verified the occurrence of many estimation outliers in the results. We defined an estimation outlier as a value of the dependent variable $y$ with a percentage error more significant than 10% relative to its respective observed sample. The percentage error ($e_{pi}$) of a sample of index "$i$" is defined by

$$e_{pi} = 100 \cdot \frac{|\hat{y}_i - y_i|}{y_i}, \qquad (1)$$

where: $y_i$ is the observed value of the dependent variable of the sample $i$, and $\hat{y}_i$ is the estimated value of the dependent variable relative to the sample $i$.

Upon examining the results derived from the EZ models, we noted that most outliers occurred when the aspect angle absolute value was equal to or exceeded 144°. Consequently, we elected to generate additional maximum range samples for the two SAMs within the 144° to 180° range to reduce the number of outliers in the testing phase of the EZ models. To achieve this, we partitioned the domain of the aspect angle modulus into five distinct intervals or sectors: [0°, 144°), [144°, 153°), [153°, 162°), [162°, 171°), and [171°, 180°). We then generated 5000 samples within each interval, utilizing LHS for this purpose. The experiments were executed using an Intel Xeon Gold 6230R CPU with 2.10GHz (4GHz of maximum turbo frequency) and 64 GB of RAM.

Finally, we created an EZ model for each sample set of the SAM$_1$ and SAM$_2$ through the application of methods based on Artificial Neural Network (ANN), Random Forest Regressor (RFR), and Polynomial Regression (PR). Additionally, we developed comprehensive EZ models that incorporated the entirety of the sample set – essentially, the 25,000 samples spanning the interval [0°, 180°] for each SAM type.

When developing the EZ models for each sample set, we made the following considerations:

- We allocated 80% of the samples for training and validation purposes of the models;
- We employed the 5-fold cross-validation method for model training and validation; and
- The remaining 20% of the samples were reserved for testing the models.

In assessing the efficacy of the EZ models, our primary goal was to identify the model that exhibited the lowest mean percentage error. This measure would suggest a reduced incidence of outliers. We adopted a collection of metrics to examine these models, as detailed in the results section of our report. This in-depth analysis helped us determine whether using a singular EZ model for the entire interval of [0°,180°] would be more beneficial or if dividing models for each of the five sectors might yield better outcomes. Additionally, this approach allowed us to compare the models derived from previously mentioned supervised learning techniques: ANN, RFR, and PR.

Tables 1 and 2 offer an in-depth overview of the critical hyperparameters used in the ANN and RFR methodologies for constructing the EZ models, respectively. These configurations were meticulously chosen through exhaustive experimentation. We utilized the TensorFlow library (Abadi et al., 2016) for the ANN methodology, with each hidden layer having a uniform number of units. For the RFR methodology, we employed the Scikit-Learn library (Pedregosa et al., 2011).

**Table 1. Detailed settings of key hyperparameters for the ANN model.**

| Hyperparameter | Values |
|---|---|
| Number of Hidden Layers | [2, 5, 10] |
| Number of Units in Each Hidden Layer | [32, 64, 128] |
| Batch Size | 16 |
| Patience Threshold for Early Stopping | 10 |
| Optimizer | Adam* |
| Activation Function in the Hidden Layers | Rectified Linear Unit (ReLU) |

*Optimizer based on the method proposed by Kingma and Ba (2014).

**Table 2. Detailed settings of key hyperparameters for the RFR model.**

| Hyperparameter | Values |
|---|---|
| Number of Estimators | 100 |
| Maximum Depth of a Tree | None (Unrestricted) |
| Minimum Number of Samples to Split a Node | 2 |
| Minimum Number of Samples at a Leaf Node | 1 |
| Maximum Number of Features for Optimal Split | 3 |
| Bootstrap | True |

Our study explored other machine learning techniques, such as the Support Vector Machine and AdaBoost. However, the results from these methods significantly underperformed compared to those from the ANN and RFR. As a result, we chose not to include these underperforming methods in this paper to maintain conciseness and to highlight the more promising approaches. We used the "polyFit" function from the R library to establish the PR model. Table 3 illustrates the hyperparameter configurations for this function. Additional parameters related to the ANN, RFR, and PR methodologies were left at their default settings, and we opted not to detail them further.

**Table 3. Detailed settings of key hyperparameters for the PR model.**

| Hyperparameter | Values |
| --- | --- |
| Maximum Degree for Polynomial Terms | [9, 10, 12, 13, 14, 15] |
| Maximum Degree of Interaction Terms | 3 |

The results of creating these EZ models from the 12 sample sets and comparing them are described in the next section.

**RESULTS**

Each SAM system, specifically $SAM_1$ and $SAM_2$, incorporates five distinct sample sets, as previously mentioned. Additionally, we included a sixth sample set, which is functionally identical to the collective union of the original five sample sets, covering an interval range from 0° to 180°. This expanded dataset allowed us to construct 36 predictive models, evenly split with 18 dedicated to each SAM type.

The model development process utilized the already mentioned methodologies, resulting in an assortment of six models for each method, amounting to 18 models per SAM system.

We assessed these models using the following performance metrics:

- Coefficient of Determination ($R^2$);
- Root Mean Squared Error (RMSE) expressed in nautical miles (nm);
- Mean Absolute Percentage Error (MAPE); and
- Processing Time (PT) measured in seconds.

The evaluation process of these models consisted of training and testing phases. Detailed information about these phases is provided in the subsequent subsections.

**Training Phase**

We analyzed and compared the performance of the training and validation phases of the EZ models by applying the 5-fold cross-validation method to 80% of their respective sample sets. We did a comparative analysis among the EZ models created by the same method.

We verified that the PR method created EZ models that performed better for the sectors than for the whole interval [0°, 180°], considering the $SAM_1$. The worst performance of PR was in the sector [162°, 171°], presenting an RMSE value of 1.4429 nm ± 0.1765 nm and a MAPE value of 2.85% ± 0.23%. For the interval [0°, 180°], PR showed an RMSE value of 2.0775 nm ± 0.0331 nm and a MAPE value of 8.54% ± 0.31%. Considering the $SAM_2$, we verified that the PR method created EZ models that also performed better for the sectors than for the whole interval [0°, 180°], except the sector [0°, 144°]. PR presented an RMSE value of 3.5196 nm ± 0.1703 nm and a MAPE value of 14.05% ± 2.42% for this sector. Whereas for the whole interval, PR showed an RMSE value of 2.8757 nm ± 0.1688 nm and a MAPE value of 5.09% ± 0.75%.

Regarding the ANN models for the $SAM_1$, the model performance was notably superior in the [162°, 171°] sector (5 hidden layers with 64 units), with the highest mean $R^2$ value of 0.9986 ± 0.0005 and the smallest mean RMSE of 0.5113 nm ± 0.0996 nm and MAPE as 1.71% ± 0.34%, suggesting reasonable predictions. The sector [0°, 144°] (10 hidden layers with 32 units) also showed promising results, recording an $R^2$ of 0.9931 ± 0.0034 and a relatively low RMSE and MAPE, 0.2630 nm ± 0.0645 nm and 3.79% ± 1.15% respectively. However, the performance significantly diminished for the more comprehensive sector range, especially in the [0°, 180°] sector (2 hidden layers with 32 units), where it had the lowest $R^2$ of 0.9718 ± 0.0280 and the highest RMSE of 2.061 nm ± 1.157 nm and MAPE of 7.76% ± 1.95%, indicating a slight deterioration in the prediction process. Among the $SAM_2$ model sectors, the highest performing sector was the [162°, 171°] range (5 hidden layers with 64 units), with a remarkable mean $R^2$ value of 0.9991 ± 0.0004 and a significantly lower RMSE of 0.2274 nm ± 0.0541 nm and MAPE of 0.24% ± 0.06%. The sector from [171°, 180°] (5 hidden layers with 32 units) also showed strong performance with a mean $R^2$ value of 0.9975 ± 0.0016 and low RMSE and MAPE values. Conversely, the [144°, 153°] (5 hidden layers with 64 units) exhibits a slightly lower mean $R^2$ value of 0.9644 ± 0.0274 and an RMSE of 2.0155 nm ± 0.7836 nm and MAPE of 1.64% ± 0.81%. The broadest sector, spanning [0°, 180°] (2 hidden layers with 128 units), posts a mean $R^2$ of 0.9942 ± 0.0020 and an RMSE of 1.7500 nm ± 0.2713 nm, indicating good overall performance. As a general trend, it can be inferred that models trained on narrower sector ranges tend to yield superior performance.

For $SAM_1$, we observed that RFR created EZ models that performed better for the whole interval, except for sectors [0°, 144°) and [144°, 153°). For the first sector, RFR presented an RMSE value of 0.2452 nm ± 0.0243 nm and a MAPE value of 3.22% ± 0.15%. For the second sector, RFR showed an RMSE value of 0.3462 nm ± 0.0687 nm and a MAPE value of 1.52% ± 0.60%. Whereas for the whole interval, RFR showed an RMSE value of 0.6197 nm ± 0.0668 nm and a MAPE value of 1.71% ± 0.14%. We also verified that, regarding MAPE values, the EZ model created for the whole interval performed better than the EZ model designed for the first interval. For $SAM_2$, we observed that RFR created EZ models that performed better for the sectors, except for sectors [0°, 144°) and [144°, 153°). For the first sector, RFR returned an RMSE value of 2.9785 nm ± 0.3371 nm and a MAPE of 6.75% ± 2.33%. For the second sector, RFR presented an RMSE value of 2.5176 nm ± 0.4481 nm and a MAPE value of 1.25% ±0.24%. Whereas for the whole interval, RFR returned an RMSE value of 1.8977 nm ± 0.1755 nm and a MAPE value of 2.09% ± 0.53%. Regarding MAPE values, we also perceived that EZ models built for the second sector performed better than the EZ model of the whole interval.

**Testing Phase**

In this phase, we tested the EZ models using new sample sets not utilized during training. Subsequently, we analyzed and compared the performance of the EZ models using the remaining 20% of their respective sample sets. The metric results are presented in Table 4 for the $SAM_1$ and Table 5 for the $SAM_2$.

**Table 4. Evaluation of EZ models for $SAM_1$ during the testing phase using 20% of the sample sets.**

| [0°, 144°) | PR | ANN | RFR | [144°, 153°) | PR | ANN | RFR |
|---|---|---|---|---|---|---|---|
| $R^2$ | 0.9996 | 0.9979 | 0.9949 | $R^2$ | 0.9939 | 0.9940 | 0.9919 |
| RMSE | 0.0646 | 0.1578 | 0.2278 | RMSE | 0.2369 | 0.2288 | 0.2740 |
| MAPE | 0.92% | 2.23% | 3.32% | MAPE | 0.82% | 1.21% | 0.98% |
| PT | 0.5070 | 0.2465 | 0.0255 | PT | 1.8970 | 0.1926 | 0.0247 |
| [153°, 162°) | PR | ANN | RFR | [162°, 171°) | PR | ANN | RFR |
| $R^2$ | 0.9881 | 0.9944 | 0.9927 | $R^2$ | 0.9873 | 0.9980 | 0.9947 |
| RMSE | 0.6578 | 0.4398 | 0.5333 | RMSE | 1.4296 | 0.5523 | 0.9022 |
| MAPE | 1.62% | 1.24% | 1.25% | MAPE | 2.93% | 0.09% | 1.44% |
| PT | 1.9120 | 0.2661 | 0.0251 | PT | 1.9360 | 0.1812 | 0.0256 |

| [171°, 180°] | PR | ANN | RFR | [0°, 180°] | PR | ANN | RFR |
|---|---|---|---|---|---|---|---|
| $R^2$ | 0.9942 | 0.9976 | 0.9976 | $R^2$ | 0.9764 | 0.9914 | 0.9389 |
| RMSE | 1.0533 | 0.6713 | 0.6779 | RMSE | 2.0538 | 1.2869 | 3.4020 |
| MAPE | 2.41% | 1.50% | 0.96% | MAPE | 8.04% | 7.27% | 4.39% |
| PT | 1.3440 | 0.1771 | 0.0255 | PT | 8.6560 | 0.2763 | 0.0778 |

**Table 5. Evaluation of EZ models for $SAM_2$ during the testing phase using 20% of the sample sets.**

| [0°, 144°) | PR | ANN | RFR | [144°, 153°) | PR | ANN | RFR |
|---|---|---|---|---|---|---|---|
| $R^2$ | 0.9766 | 0.9800 | 0.9809 | $R^2$ | 0.9583 | 0.9823 | 0.9675 |
| RMSE | 3.1722 | 3.1006 | 2.9824 | RMSE | 2.2741 | 1.5595 | 1.9888 |
| MAPE | 14.51% | 14.21% | 6.08% | MAPE | 1.57% | 0.98% | 0.99% |
| PT | 0.5170 | 0.1328 | 0.0267 | PT | 2.9570 | 0.1836 | 0.0266 |
| [153°, 162°) | PR | ANN | RFR | [162°, 171°) | PR | ANN | RFR |
| $R^2$ | 0.9989 | 0.9778 | 0.9788 | $R^2$ | 0.9999 | 0.9988 | 0.9995 |
| RMSE | 0.2432 | 1.1425 | 1.1029 | RMSE | 0.0358 | 0.2695 | 0.1655 |
| MAPE | 0.17% | 0.37% | 0.30% | MAPE | 0.04% | 0.29% | 0.15% |
| PT | 0.7480 | 0.2586 | 0.0262 | PT | 0.5590 | 0.1843 | 0.0260 |
| [171°, 180°] | PR | ANN | RFR | [0°, 180°] | PR | ANN | RFR |
| $R^2$ | 0.9999 | 0.9991 | 0.9997 | $R^2$ | 0.9849 | 0.9836 | 0.9993 |
| RMSE | 0.0418 | 0.2251 | 0.1401 | RMSE | 2.8116 | 3.001 | 0.2078 |
| MAPE | 0.04% | 0.23% | 0.13% | MAPE | 4.63% | 5.78% | 0.21% |
| PT | 0.7450 | 0.1954 | 0.0255 | PT | 12.9340 | 0.2693 | 0.1027 |

Regarding $R^2$, ANN and RFR consistently outperform PR across most sections in both $SAM_1$ and $SAM_2$. The highest $R^2$ observed for ANN was 0.9991 in $SAM_2$, and for RFR was 0.9997 in $SAM_2$. In contrast, PR peaked at 0.9996 in $SAM_1$ and 0.9999 in $SAM_2$ but frequently exhibited lower $R^2$ values than the other two models.

For RMSE, ANN performs better in $SAM_1$, while PR and RFR perform better in $SAM_2$.

The MAPE comparison reveals a mixed picture. PR shows the highest error in $SAM_1$ at 8.04% but performs well in $SAM_2$ with the lowest MAPE of 0.04%. ANN's performance varies, with the highest error observed at 14.21% in $SAM_2$ and the lowest at 0.09% in $SAM_1$. RFR consistently performs with lower errors, the best being 0.13% in $SAM_2$. PT shows a clear trend: both ANN and RFR significantly outperform PR. RFR consistently shows the fastest processing time across all sections in both $SAM_1$ and $SAM_2$, indicating its computational efficiency. Through the results presented in Tables 4 and 5, we found that the estimation times of all models are lower than the threshold of 0.01 seconds ($L$). The time to estimate the maximum engagement range of a single missile shot is equal to PT divided by the number of samples used in the testing phase of the respective model. For example, in Table 5, using the same hardware, the estimation time of the EZ model, created by the PR method for the interval [0º, 180º], is 12.9340 ÷ 5000 = 0.0026 seconds. This time limit, $L$, is significantly lower than the estimation time of the simulation tool. The simulation tool takes around 34 seconds to estimate the maximum engagement range of a single missile shot.

We also did a comparative analysis among the EZ models created by the same method. We observed that the EZ models produced by the PR method presented performances similar to those obtained by the EZ models developed in the training phase. The performance trend of the EZ models created by the ANN method is similar to that of the EZ models produced by the PR method. For $SAM_1$, the EZ models made by the ANN method performed better for the sectors than for the whole interval. For $SAM_2$, EZ models built by the ANN method also performed better for the sectors than for the entire interval, except for the sector [0°, 144°). We noted different performances related to the RFR method. For $SAM_1$, EZ models built by the RFR method performed better for the sectors than for the whole interval. For $SAM_2$, EZ models built by the RFR method performed better for the whole interval, except sectors [162°, 171°) and [171°, 180°]. However, the differences between the RMSE value returned by the EZ model of the whole interval and the RMSE values of these sectors are lower than 0.2 nm. Similarly, the differences between the MAPE value of the whole interval and the MAPE values of these sectors are lower than 0.1%. These differences' magnitude is insignificant in the context of the problem addressed in this work. Therefore, we considered that the performance of the EZ model of the whole interval is equivalent to the performance of the EZ models of these two sectors.

Finally, we did a comparative analysis between the methods, i.e., we compared EZ models created by one technique with EZ models created by another.

We understood that the best composition of a SAM's EZ depends on the simulation work's purpose. Therefore, we could represent an EZ of a SAM through the following:

- A Single EZ Model: This involves using one individual model to represent the SAM's EZ;
- A Homogeneous Multimodel: This refers to a collection of EZ models created by the same method; and
- A Heterogeneous Multimodel: In contrast, this encompasses a set of EZ models produced by different methods.

If we only prioritized the most accurate EZ model, i.e., the performance in terms of the RMSE and MAPE values, we could represent the EZ for the $SAM_1$ as a heterogeneous multimodel. This multimodel would be formed by an EZ model built using the PR method for the sector [0°, 144°) and EZ models created using the ANN method for the remaining sectors [144º, 153º), [153°, 162°), [162°, 171°) and [171°, 180°]. Conversely, the EZ for the $SAM_2$ could be represented by a single EZ model built using the RFR method for the interval [0°, 180°], with an RMSE value of 0.2078 and a MAPE value of 0.21%. The EZ models built using the PR method for the sectors [162°, 171°) and [171°, 180°] presented RMSE and MAPE values lower than 0.2078 and 0.21%, respectively. However, the differences between these RMSE and MAPE values and the mentioned values are lower than 0.2 nm and 0.1%. As we explained previously, these values are insignificant in comparing methods. Thus, we considered that the EZ models built by the PR method for these two sectors are equivalent to the EZ model produced by the RFR method for the whole interval [0°, 180°].

Suppose we prioritized the fastest EZ model, i.e., the performance in terms of the estimation time, with an RMSE limit of around 1.0 nm. In that scenario, we could represent the EZ for the $SAM_1$ as a homogeneous multimodel. Models constructed using the RFM method for the five sectors [0º, 144°), [144º, 153º), [153°, 162°), [162°, 171°), and [171°, 180°] would form this multimodel. The EZ for the $SAM_2$ could be represented as a single EZ model for the whole interval [0°, 180°].

In the context of this work, all three models exhibit strengths and weaknesses. In that case, PR tends to show the least prediction error but falls short in processing time efficiency. The RFR stands out for its processing time efficiency and presents a strong performance in error reduction. ANN also delivers a balanced performance, doing well in error reduction but lagging behind RFR in processing speed.

**CONCLUSION AND FUTURE WORK**

This study has illuminated the transformative potential of machine learning techniques in improving the efficiency and speed of SAM EZ simulations. The conventional computational limitations, which have consistently interfered with defense strategy planning, can be mitigated significantly by implementing machine learning algorithms. The integration of machine learning and custom-designed simulation tools within this study has yielded promising results, demonstrating the feasibility of predicting SAM EZs with precision and efficiency.

Our analysis of three machine learning models produced informative findings. PR showed the least prediction error, but its processing time efficiency could be improved. On the other hand, RFR brought its superior processing time efficiency and demonstrated commendable performance in error reduction. ANN also showed satisfactory performance in reducing error, although it lagged behind RFR in processing speed.

Despite the encouraging results, our study recognizes the inherent challenges in applying machine learning techniques in this context, such as the need for extensive training data, the risk of model overfitting, and the continual necessity for meticulous evaluation and refinement to maintain accuracy. Yet, these challenges provide pathways for future research.

Future work directions include further optimizing these machine-learning models and reducing processing time while maintaining or improving prediction accuracy. Moreover, researchers should explore incorporating other machine learning models not covered in this study. Considering the fast-paced evolution of algorithms, we anticipate novel methodologies may yield even more promising results in predicting SAM EZs. Also, a significant area of future research should include strategies to manage the challenges presented by data requirements and model overfitting, such as techniques for efficient data augmentation and model regularization.

In conclusion, our work affirms the potential of machine learning in advancing air defense strategies, offering a new level of dynamism and responsiveness to military decision-making. While the path toward full implementation of these techniques in the defense sector is not without its challenges, this study underscores the promise held by this field, paving the way for future innovative solutions.

**SOURCE CODE**

For further insight and access to the source code of this research, please visit our repository at https://github.com/jpadantas/sam-ez.

**ACKNOWLEDGEMENTS**

This work has been supported by Finep (Grant nº 2824/20). Takashi Yoneyama and Marcos R. O. A. Maximo are partially funded by CNPq – National Research Council of Brazil – through grants 304134/2-18-0 and 307525/2022-8, respectively.